\documentclass{article}

\usepackage[utf8]{inputenc} 
\usepackage[T1]{fontenc}    
\usepackage{hyperref}       
\usepackage{url}            
\usepackage{booktabs}       
\usepackage{amsfonts}       
\usepackage{nicefrac}       
\usepackage{microtype}      

\usepackage{amsmath}
\usepackage{amssymb}

\usepackage{graphicx}
\usepackage{subfigure}
\usepackage{natbib}
\usepackage{fullpage}

\graphicspath{{./}}

\usepackage{lmodern}

\title{
Learning and generalization of compositional descriptions of visual scenes
}

\usepackage{authblk}
\author[1,2]{E. Paxon Frady}
\author[2]{Spencer Kent}
\author[2]{Quinn Tran} 
\author[2]{Pentti Kanerva}
\author[2]{Bruno A. Olshausen}
\author[1,2]{Friedrich T. Sommer}

\affil[1]{Intel Labs}
\affil[2]{Redwood Center for Theoretical Neuroscience, UC Berkeley}

\begin{document}

\maketitle

\begin{abstract}
Complex visual scenes that are composed of multiple objects, each with attributes, such as object name, location, pose, color, etc., are challenging to describe in order to train neural networks.
Usually, deep learning networks are trained supervised by categorical scene descriptions.
The common categorical description of a scene contains the names of individual objects but lacks information about other attributes. 
Here, we use distributed representations of object attributes and vector operations in a vector symbolic architecture to create a full compositional description of a scene in a high-dimensional vector.  
To control the scene composition, we use artificial images composed of multiple, translated and colored MNIST digits.
In contrast to learning category labels, here we train deep neural networks to output the full compositional vector description of an input image. 
The output of the deep network can then be interpreted by a VSA resonator network, to extract object identity or other properties of indiviual objects.
We evaluate the performance and generalization properties of the system on randomly generated scenes.
Specifically, we show that the network is able to learn the task and generalize to unseen seen digit shapes and scene configurations. Further, the generalisation ability of the trained model is limited. For example, with a gap in the training data, like an object not shown in a particular image location during training, the learning does not automatically fill this gap.


\end{abstract}

\section{Introduction}
The structure of visual scenes can be described on different levels. For example, object recognition provides an abstract symbolic description of a scene, by mapping an image to category labels, describing the objects present in a scene. A limitation of object categorization is that the information about object attributes is implicit, not explicit. Other forms of scene descriptions attempt to capture the "what" and "where" by relating categoric labels to the image structure. For example, pixels (in scene segmentation) or bounding boxes (in object localization) can be assigned to different objects that compose a scene.  Here we present a model that learns to describe the semantic structure of scenes at greater detail using structured symbolic distributed representations.   

Deep learning networks trained by supervised learning with object labels reach unprecedented levels of performance in object classification when trained with large amounts of data.  Here we propose to describe a scene in terms of attributes of objects, including not only object category, but also attributes such as position, pose, and color. The scene description is represented by a high-dimensional vector using syntax rules from Vector Symbolic Architecture \citet{Plate2003, Kanerva2009}, a class of connectionist models for cognition that have been introduced into deep learning earlier \citet{Eliasmith2012}. The resulting vector is then used as training input in a deep learning network. To interpret an image, the trained network maps the input to a distributed vector representing its semantic compositional description. This vector can be queried, or all of its components disentangled one-by-one, through subsequent processing in a recurrent VSA resonator network \citep{frady_2020_resonator}
 
We perform simulation experiments on synthetic data, scenes composed of multiple MNIST digits, on how a network can be trained and queried concerning the attributes of objects in a given input image. The learning generalizes in several ways. First, the network can analyze scenes composed of MNIST digits withheld in the test set not shown during training. 
Second, the network can generalize to scenes of digits where the combination of digits shown in the scene was not shown in the training set.
Third, the network can generalize to scenes with more or fewer individual objects than what was used in the training set. 
Finally, however, the network does not generalize or transfer knowledge across the conjunction of factors that describe a scene. 
In essence, we show that if a network is not trained with a particular conjunction of factors, such as it is never shown a `7' in the `bottom-left' during training, then during testing it will not output `7' in the `bottom-left' even though it has seen 7s before and it has seen other digits in the bottom left. 

\section{Background}
Based on frameworks of computing with high-dimensional vectors \citep{Plate2003, Kanerva2009}, our group has recently made progress to enable neural networks to learn and manipulate compositional structure. 
Our recent theoretical work \citep{frady_2018sequence} shows that several frameworks, collectively called \emph{Vector Symbolic Architectures} (VSA) \citep{Gayler2003}, are mathematically related and have universal properties, such as working memory capacity.
VSAs are formal, comprehensible frameworks for representing information and computing with random high-dimensional vectors, akin to neural population activity. The frameworks include operations, such as superposition (e.g. add), variable binding (e.g. componentwise multiply) and permutation of coordinates, that enables the expression of compositional structure in a distributed vector representation. 
Further, VSAs can be used to express symbolic computations in a neural network, which allows integration of deep learning networks to create cognitive agents \citep{Eliasmith2012}. The essential properties of VSAs are:
\begin{itemize}
\item
Symbols and compositions are represented with high-dimensional vectors (neural population activity). 
\item
Elementary symbols are chosen randomly and stored in a codebook (matrix of synaptic weights). In high-dimensions, random vectors are close to orthogonal with high probability.
\item
Multiple symbols can be represented in superposition through vector addition. However, superposition is limited by crosstalk noise.
\item
The binding operation forms compositions of symbols through vector multiplication. This operation is invertible. However, given a pair of bound vectors, one is needed to extract the other.
\end{itemize}

The issue of what and where has been heavily studied in the machine-learning literature. Approaches include forming bounding-boxes around objects or classifying each pixel as part of an object. While these approaches have shown some success in dealing with natural images, it is not fully clear how to extend them further. 


Compositions are critically important to VSAs.  However, a fundamental problem arises if multiple components of a composition are unknown. The typical solution requires enumerating the entire set of combinations, which leads to problems with scaling. 



To solve such combinatoric inference problems, we have developed a recurrent neural network architecture called the \emph{resonator network} based on VSA principles. 
The VSA resonator network is related to and generalizes map-seeking circuits \citep{Arathorn2002} and dynamic routing \citep{Olshausen1995}. It iterates and converges to the best solution by using the principle of superposition to search over many guess combinations in parallel \citep{Frady2020}.


\section{Methods}
\subsection{Structured compositional vectors for describing complex scenes}
\begin{figure}
    \centering
    \includegraphics[width=\textwidth]{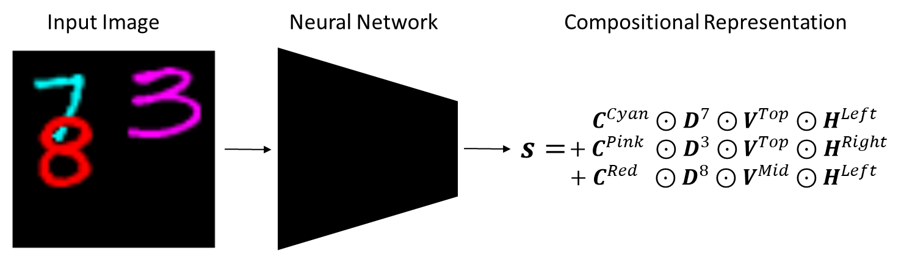}
    \caption{A scene is generated with 1-3 random MNIST digits with independent colors and locations.A multi-layer feed-forward network was trained to directly transform an input image into a structured compositional vector representation of the scene. The compositional representation can be used for reasoning, transforms and scene understanding. However, the components are still entangled and need to be inferred. }
    \label{fig:comp_learn}
\end{figure}

To describe a complex scene, such as the one shown in Figure \ref{fig:comp_learn} (left panel), the VSA frameworks provide two essential operations: binding and superposition. The binding operation, elementwise multiply ($\odot$), can attach symbols together into a compound representation, allowing one to associate properties to the same object. Superposition, vector addition ($+$), allows for multiple objects to be simultaneously represented. 

Each individual component can be enumerated and described by a random base vector. The set of base vectors describing all possible properties within a particular class is stored in a matrix called the \emph{codebook}. In this case, there are 7 colors, 10 numbers, 3 y-positions, and 3 x-positions, giving 630 combinations. These have corresponding codebooks of randomly chosen -1 or +1: $\mathbf{C} : N \times 7$, $\mathbf{D} : N \times 10$, $\mathbf{V} : N \times 3$, and $\mathbf{H} : N \times 3$, where $N=1000$ is the number of neurons. 

Each object in a complex scene can then be described with a compound vector. The vector for color, digit identity, and location are bound together through multiplication to create the compound vector. The scene vector $\mathbf{s}$ is given by summing the compound vector for each object, and acts as a compositional representation for the entire scene:
\begin{equation}
    \mathbf{s} = \sum_{i}^{objects} \mathbf{C}_{color_i} \odot \mathbf{D}_{digit_i} \odot \mathbf{V}_{y-pos_i} \odot \mathbf{H}_{x-pos_i}
    \label{eq:scene}
\end{equation}

\subsection{Learning to output structured vector representations}
We demonstrate the resonator in conjunction with a neural network that learns to assign compound vectors to images of simple scenes. 
The neural network is a multi-layer perceptron with two hidden layers.
A scene is generated by choosing a random combination of colors, MNIST digits, and locations, and the corresponding compound vector acts as a description of the scene.
Through supervised training, the network learns to map the generated image into the compound vector that describes the scene. 

The output of the network is a compound distributed representation of the scene, which the VSA resonator network can factor into its components. 
Due to correlations and ambiguity in the space of images, the network may not be able to produce precisely the ground truth compound vector. However, the nature of distributed vectors in vector symbolic architectures allows for noise tolerance.

\subsection{Factoring compositional vectors with resonator networks}
In order to determine the components of a compound vector, we must search for combinations of vectors in the codebooks that produce a vector similar to our compound vector. If we were using an associative memory, such as a Hopfield Network
or a Sparse Distributed Memory, to do the search, then we would have to store all possible combinations of vectors in the memory. This in many cases would be impractical. 

Our algorithm allows one to store only the codebooks, but still factor compound vectors. To infer one of the components of a compound vector, we use the principle of superposition to simultaneously check multiple guesses. For each component class, we instantiate a \emph{resonator module}: $\mathbf{\hat{c}}(t)$, $\mathbf{\hat{d}}(t)$, $\mathbf{\hat{v}}(t)$, and $\mathbf{\hat{h}}(t)$, which are high-dimensional vectors that act as an estimate of the factor's parameters. The resonator modules are connected into a network based on the generative model of the scene $\mathbf{s}$ (\ref{eq:scene}). The full dynamics for VSA resonator network for such scenes is given by:
\begin{align}
\begin{split}
\mathbf{\hat{c}}(t+1) &= f \left( \mathbf{C}\mathbf{C}^{\top} \left( \mathbf{s} \odot \mathbf{\hat{d}}(t) \odot \mathbf{\hat{v}}(t) \odot \mathbf{\hat{h}}(t) \right) \right) \\
\mathbf{\hat{d}}(t+1) &= f \left( \mathbf{D}\mathbf{D}^{\top} \left( \mathbf{s} \odot \mathbf{\hat{c}}(t) \odot \mathbf{\hat{v}}(t) \odot \mathbf{\hat{h}}(t) \right) \right) \\
\mathbf{\hat{v}}(t+1) &= f \left( \mathbf{V}\mathbf{V}^{\top} \left( \mathbf{s} \odot \mathbf{\hat{d}}(t) \odot \mathbf{\hat{c}}(t) \odot \mathbf{\hat{h}}(t) \right) \right) \\
\mathbf{\hat{h}}(t+1) &= f \left( \mathbf{H}\mathbf{H}^{\top} \left( \mathbf{s} \odot \mathbf{\hat{d}}(t) \odot \mathbf{\hat{v}}(t) \odot \mathbf{\hat{c}}(t) \right) \right) \\
\end{split}
\label{eqn:resonator}
\end{align}
where $f$ is the sign function or normalization. 

\begin{figure}
    \centering
    \includegraphics[width=0.9\textwidth]{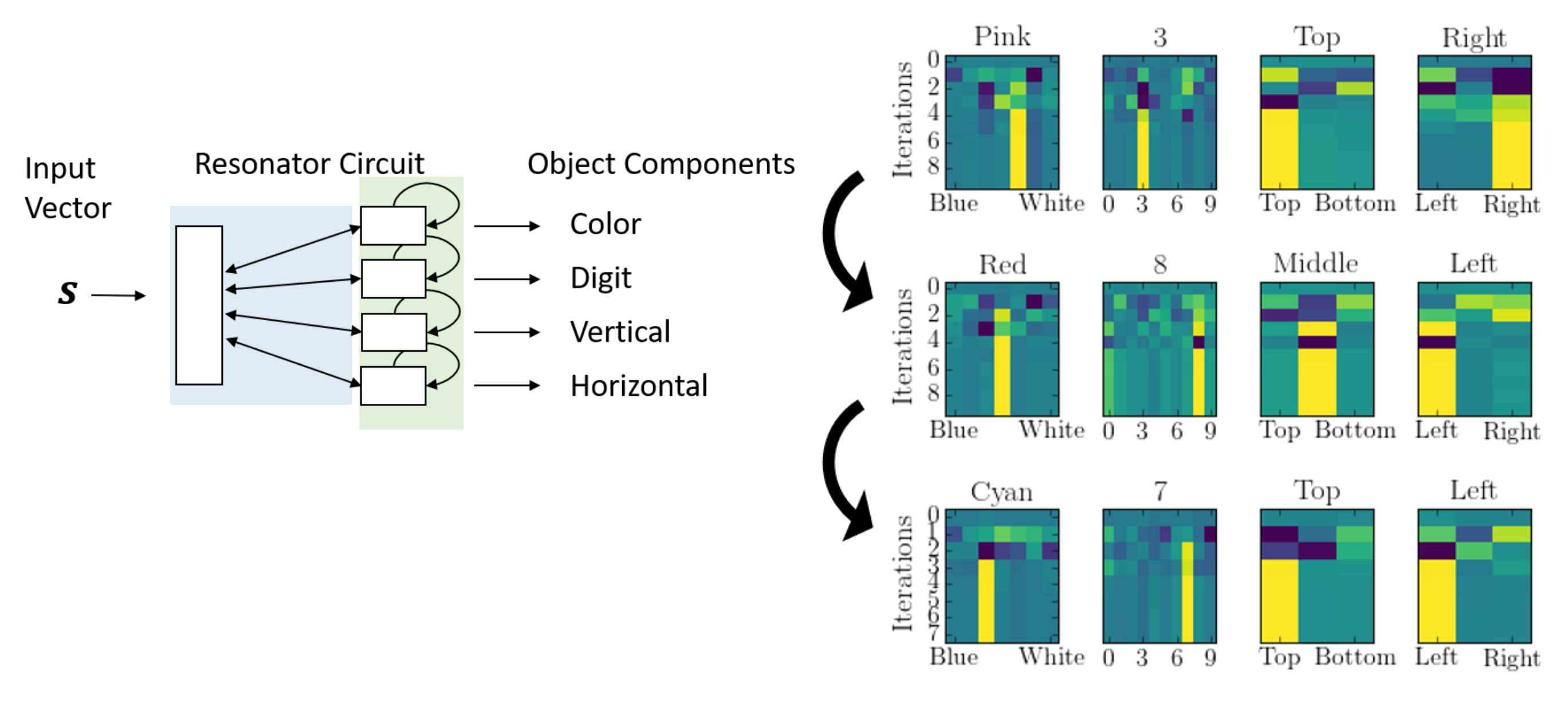}
    \caption{The output of the deep network $\mathbf{s}$ is input into the VSA resonator network, and the system iterates until convergence. Each resonator module can represent multiple guesses simultaneously. Each module relies on the guesses from other modules to try and infer the correct component (blue box). The resonator modules then \emph{clean-up} their individual guesses to produce a smaller range of guesses in the next iteration. The resonator network converges to a solution that describes the components of one of the objects in the scene. The state of each resonator module each time-step is visualized as a heatmap, each row is an iteration and each column is one possible component. The component with the largest weight is chosen as the output (top of each panel; yellow). The object is then explained away by subtracting the networks’s guess from the scene vector, $\mathbf{s}$. The resonator network is reset and converges to another solution, which describes a different object in the scene. This is repeated until all of the objects in the scene are recognized.
}
    \label{fig:res_circuit}
\end{figure}

The VSA resonator network can be visualized in Figure \ref{fig:res_circuit}. Each resonator module can be initialized randomly. The activity state of each resonator represents a set of guesses for the particular component. Superposition allows for multiple guesses to be represented simultaneously in a high-dimensional vector, but crosstalk noise increases with more guesses in superposition. The guesses from the other resonators are used to infer the component from the scene vector, $\mathbf{s}$ (Fig. \ref{fig:res_circuit} blue box). After the inference step, the guesses are \emph{cleaned-up} by the codebooks (Fig. \ref{fig:res_circuit} green box). This provides better guesses each iteration, which in turn reduces the effect of crosstalk noise. The correct decomposition of the compound vector then emerges through a positive feedback cycle -- the correct solution \emph{resonates}, and the system converges to a stable point.
Once a stable point is reached, the network outputs can be decoded into the properties of an individual object in the scene. 
Finally, to anlayze multiple objects, the previous outputs are ``explained away'' by subtracting them from the input vector $\mathbf{s}$. The network is reset and run again until convergence. This process can be repeated to deal with different numbers of objects in the scene. 
Generally once all previous outputs are explained away, a threshold on the total energy remaining in the input vector can be used to interpret whether all of the objects in the scene were detected. This can be used to halt the analysis of multiple objects, when the total number of objects is unknown.


\section{Results}
\subsection{Extracting the components of a complex scene}
We generated synthetic images for a training set, which can contain 1 to 3 MNIST digits in different locations with independent colors. These images were provided as input to a deep neural network, which is trained to output the structured vector representation of the scene, $\mathbf{s}$. 
The scene vector is then input into the resonator network, and the resonator network iterates until convergence (Fig. \ref{fig:res_circuit}).

The resonator network will randomly choose one of the objects and decompose the object into its component parts. The network converges to a solution, which describes the components of one of the objects in the scene. The output can then be \emph{explained away} from the scene vector by reconstructing the compound vector and subtracting it from $\mathbf{s}$. With the first output explained away, the resonator network is reset. It will choose another object in the scene to decompose, and converge. This process can be repeated until all of the objects in the scene are explained away. This procedure can create a description of all objects in the scene and their components.

\subsection{Performance evaluation}
We validated performance by looking at both the output of the deep network and the output of the resonator network. 
We separated scenes with one, two or three objects, and looked at performance in each of these conditions. 

\begin{figure}
    \centering
    \includegraphics[width=0.5\textwidth]{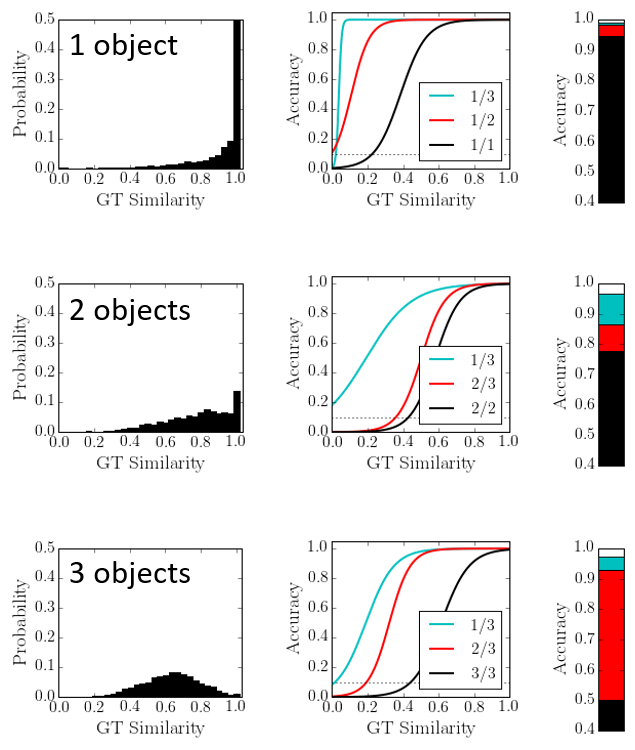}
    \caption{The output of the deep network is compared to the ground-truth scene vector based on its cosine similarity (GT Similarity) for all images in the test set (left panels). The output of the resonator is accurate if all four extracted components correctly match one of the objects in the scene. The performance of the resonator is compared conditional on the quality of the output in the deep network (middle panels). We run the resonator three times. The black lines show the system correctly identifying all objects in the scene in as few attempts as necessary. The red and cyan lines show imperfect performance. Legend: digits correct / number of runs. The overall performance is shown in the right panels.
}
    \label{fig:performance}
\end{figure}

The deep net's output was compared to the ground truth vector for the scene, this is the ``GT Similarity'' (Fig. \ref{fig:performance} left panels). The closer the output is to 1, the better the performance of the deep network.

The resonator's output was evaluated as correct if all four components (color, digit, x-location, y-location) matched one of the ground truth objects in the scene. If even one component was incorrect, then the output would be considered incorrect. Note that in these settings, if the ground truth vector for one object was given as input to the resonator network, it will extract the correct identity and other properties virtually 100\% of the time, as the network is well within the regime of its operational capacity \citep{kent_2020resonator}.

In our evaluation experiment for every input scene, we ran the resonator network three times, and subtracted the output from the scene vector between each reset. We ran three times regardless of the number of objects in the scene, however the number of objects can be easily inferred based on the output of the deep network. This information could be used to decide how many times the resonator should be reset. In these experiments, running the resonator network three times when there is only one object allows the system to make three guesses about the object. 

We evaluated the performance of the resonator network conditional on the output of the deep network (Fig. \ref{fig:performance} middle panels), as well as the overall performance of the full system (Fig. \ref{fig:performance} right panels). If the deep network output is close to the ground truth, then the resonator network is more likely to be correct. 
We find that our algorithm is able to correctly infer the parameters of these scenes even when the vector-symbolic representations are quite far from the ground truth.

When there is only one object in the scene, the performance of the entire system is above $90\%$ (Fig. \ref{fig:performance} top, black), comparable to standard deep network performance on MNIST. When there are two objects, the VSA resonator network guesses both objects correctly nearly $80\%$ of the time in the first two runs (Fig. \ref{fig:performance} middle, black). With three objects in the scene, the system accurately extracts all three objects about $50\%$ of the time in three runs (Fig. \ref{fig:performance} bottom, black), and gets at least two out of three over $90\%$ of the time (Fig. \ref{fig:performance} bottom, red). Virtually all errors are due to incorrect digit classification, rather than mistakes in color or location.

\subsection{Compositional transfer learning}

To better understand the nature of the compositional learning, we performed several experiments to test the outputs of the deep network under different training conditions. The essential questions relate to the combinatorics of the scene. Each object within the scene can have 9 locations, 7 colors and 10 digit identities, giving 630 combinations. There are then 630 potential scenes with a single digit. There are $396,900=630^2$ potential scenes with 2 digits, and over 250 million scenes with three digits. 

The training set contains 550 thousand examples split evenly between scenes with one, two, or three digits. The test set contains an exclusive set of hand-written digits that were not present during training. However, the combination of digit identities, colors and locations present in a scene may or may not overlap with the training set (i.e. a cyan 7 in the center right and a red 4 in the bottom left). Every scene is generated i.i.d. Because of the combinatorics, all 630 single digit combinations are present in the training set. But, only a tiny fraction of three digits scenes are explored during training, and there is virtually no overlap between the particular combination of objects present in the test and training sets. For the two digit scenes, about one third of the combinations shown in the test set were also shown in the training set. 
 
\begin{figure}
    \centering
    \includegraphics[width=0.8\textwidth]{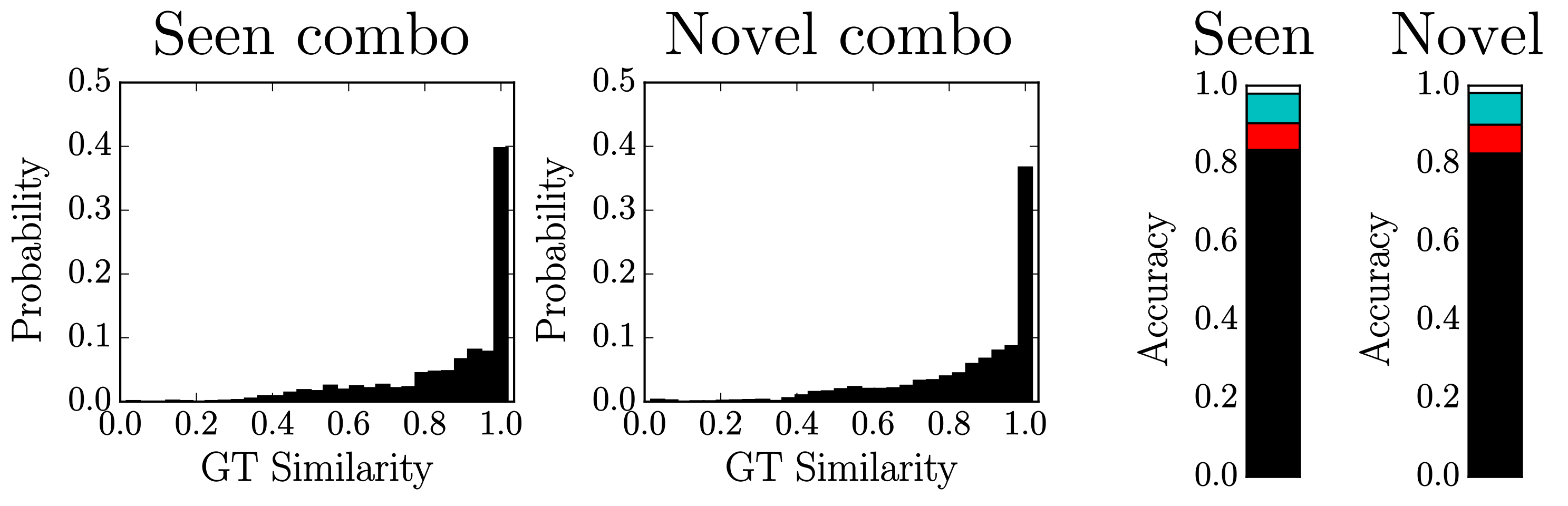}
    \caption{We isolated scenes containing the same combination of digits as shown during training, and compared them to scenes with novel combinations of digits. The performance metrics are close to indistinguishable.}
    \label{fig:combo_comparison}
\end{figure}

It is important that the network can learn across the combinatorics of each scene. Since the performance of the system on the three digit scenes is good, it is already clear that the network can handle scene combination that were not present in the training set. To see if previously seen combinations have any advantage over novel combinations, we divided scenes with two digits into two groups. The test scenes that shared the same combination of objects with a training example were separated from scenes with a novel combination of objects. Remember that the particular MNIST digit used in test scenes is always taken from a separated test set. We compared the performance of these two groups, and see little difference (Fig. \ref{fig:combo_comparison}), indicating that the combinatorics of the objects in the scene do not matter to the network. 

\begin{figure}
    \centering
    \includegraphics[width=\textwidth]{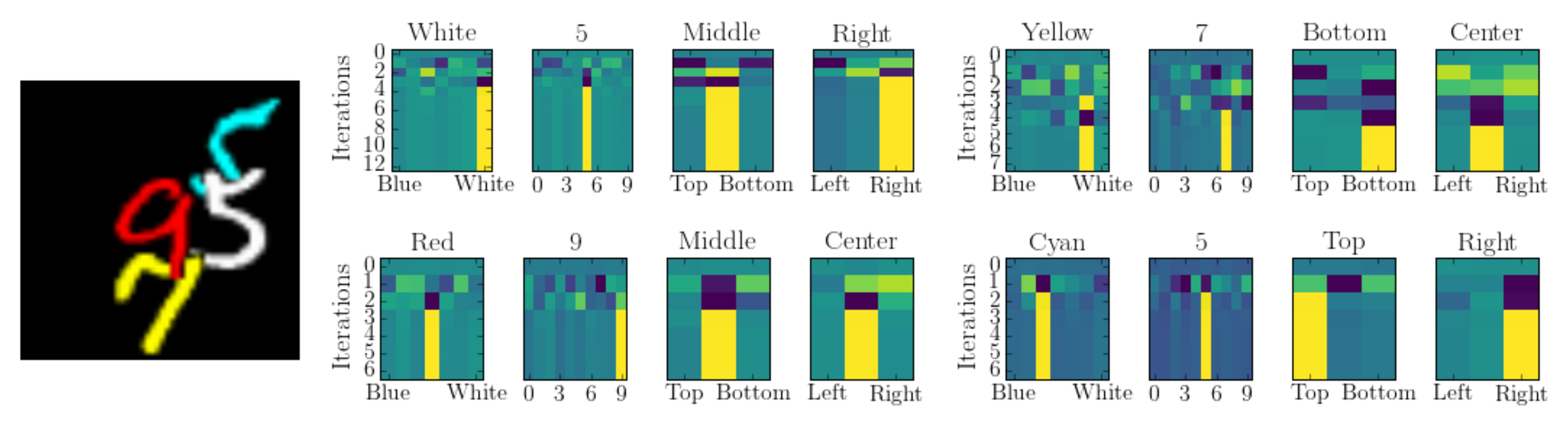}
    \caption{A network trained with maximum of three objects in a scene can successfully identify scenes with four objects.}
    \label{fig:sum_learning}
\end{figure}

The next experiment addresses whether the network can handle scenes with a different number of objects in the scene. We presented scenes with four digits to the network, and used the resonator circuit to query the output. Indeed, the system can identify four objects present in a scene, even though no scene in the training set ever contained four objects (Fig. \ref{fig:sum_learning}).

The final question is whether there is any transfer learning across the compositional properties of each object. Hypothetically, knowledge of sevens in the left can give you information of sevens on the right, but it is unclear whether the deep network can take advantage of this. The structured vectors that are used as a supervised signal contains this compositional information, but simultaneously, the structured vectors also are indicating to the network that it should treat each conjunction as unique, and that there should be no similarity between a red five and a blue five, or a red seven, or a red five in a different location. 

\begin{figure}
    \centering
    \includegraphics[width=0.6\textwidth]{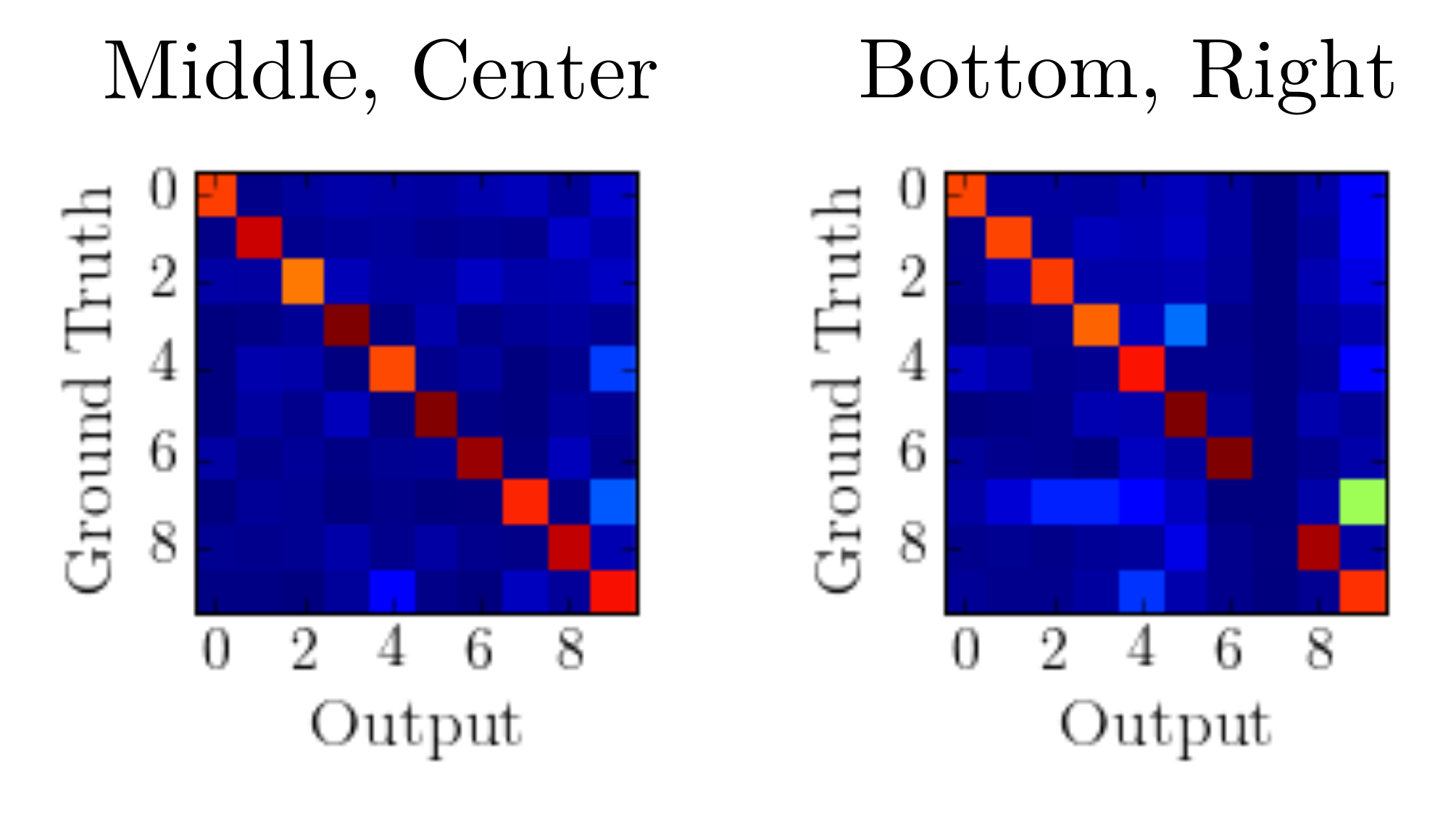}
    \caption{We removed all examples with `7' in the `bottom-right' from the training data. 
    The confusion matrix for object classification was computed for all objects in the `middle-center' location and the `bottom-right' location on the test set (which does contain `7-bottom-right').
    The network learns this gap as a property of the data, and will never output `7' in the bottom right when given the stimulus during testing.
    }
    \label{fig:transfer_learning}
\end{figure}

To see if any such transfer learning was present, we removed all `sevens' in the `bottom-right' from the training set. We then examined the outputs of the network at test time with `sevens' present in the `bottom-right'. The experiment shows that the network does not transfer knowledge about sevens from other locations to the bottom right location, but rather outputs a different digit identity (typically 9, and sometimes 2,3, or 4) (Fig. \ref{fig:transfer_learning}). 

\section{Discussion}

Based on vector symbolic architectures, here we propose a novel approach for solving scene understanding with deep learning networks. Rather than returning a list of object labels, the new approach transforms an input image into a vector description of the scene that not only contains object identities but also their poses and properties. Specifically, we show that deep networks can learn vector representations of scenes with multiple objects, formed by a vector-symbolic architecture. The scene representation is compositional, it contains information pertaining to the identity of scene objects, but also  information about their position and color. Using a VSA resonator network \citep{frady_2020_resonator}, one can extract from the output vector of the deep network the identities and properties of the individual objects. 

We analyzed simulation experiments to better understand how deep learning generalizes knowledge, learned from simple artificial scenes built from multiple MNIST digits. We find that the deep network has no issues with generalizing to digit shapes only shown during testing. We also show that the particular combination of objects present in the scene, as well as the total number of objects in the scene, do not greatly reduce the network's performance.

 We also investigated how the network can generalize over feature conjunctions. We find that if particular conjunctions of features are not present during training, then these conjunctions cannot be recognized by the deep network. For example, one might expect a network exposed to the digit shape `7' in many locations to be able to recognize a `7' in a new position. But this is not the case. If stimuli with the conjunction of `7' and `bottom-right' are removed from the training set, the network will not ever output this conjunction during testing. The network has learned this gap as a property of the data, rather than generalizing its knowledge about `7's to explain a constellation never seen during training. 

The expectation that the network generalizes knowledge of individual objects across different object poses is a typical example of inductive bias \citep{mitchell1980need}. 
Without such a generalization ability, the amount of required training data grows exponentially with number of object properties. Even when large amounts of training data are available, it is hard to track (and somehow avoid) the formation of ``gaps'' during training, in which generalization fails. Of course, for our simple artificial scenes a convolutional neural network would exhibit the required generalization ability for translation. But in more realistic scenes object variations include not only translation, but also rotation and scaling, as well as many other variable properties. Thus, a much more general form of inductive bias towards generalization is necessary. 

The inductive bias towards invariance properties is typically imposed prior to learning.  Often it involves specific network designs, like convolutional neural networks for achieving translation invariance, or networks with embedded Lie group structures to learn  transformations like translation and rotation from data \citep{chau2020}. Another popular approach is data augmentation \citep{shorten2019survey}, in which the inductive bias is imposed by adding artificially transformed data points to the training data. 

In other recent work \citep{renner2022scene}, we propose an approach to scene understanding based on analysis-by-synthesis. We use the framework of vector-symbolic-architecture with its algebra of operations for superposition and binding to formulate a generative model of scenes. 
The generative model allows us to explicitly impose the inductive bias into the resonator network to factor out invariances, such as translation, rotation, color and scaling, without any learning. 
A model of this type could be a first module before a learning deep network, where the inductive bias imposed should avoid gaps and reduce the amount of required training data. 

Ideally, however, neural networks should be able to learn from data all the transforms that objects can experience. A network should be able to learn from a video stream of the natural world how to deal with 3D geometry, transformations, and lighting conditions. Our simple experiments suggest that naive supervised learning approaches are unlikely to extract such transforms from data.

\subsubsection*{Acknowledgments}

{\small
\bibliographystyle{abbrvnat}
\bibliography{hdspike,hdprespast,capacity,references} 
}

\end{document}